\documentclass{article}

\usepackage{PRIMEarxiv}

\usepackage[utf8]{inputenc} 
\usepackage[T1]{fontenc}    
\usepackage{hyperref}       
\usepackage{url}            
\usepackage{booktabs}       
\usepackage{amsfonts}       
\usepackage{nicefrac}       
\usepackage{microtype}      
\usepackage{lipsum}
\usepackage{fancyhdr}       
\usepackage{graphicx}       
\graphicspath{{media/}}     

\usepackage{upgreek}
\usepackage{siunitx}
\usepackage{hyperref}
\usepackage{marvosym}
\usepackage[numbers,sort&compress]{natbib}
\usepackage{hyperref} 
\hypersetup{colorlinks=true,
            linkcolor=blue,
            anchorcolor=blue,
            citecolor=blue}

\title{AI-driven emergence of frequency information non-uniform distribution via THz metasurface spectrum prediction}

\author{
  Xiaohua Xing$^{1,\dag}$ Yuqi Ren$^{2,\dag}$, Die Zou$^{1}$, Qiankun Zhang$^{1}$, Bingxuan Mao$^{1}$, Jianquan Yao$^{1}$, \\ 
  \textbf{Deyi Xiong$^{2}$, Shuang Zhang$^{3}$ and Liang Wu$^{1}$} \\
 $^{1}$College of Precision Instrument and Optoelectronics Engineering, Tianjin University, \\ 
 Key Laboratory of Optoelectronics Information and Technology (Ministry of Education), Tianjin, 300072, China \\
 $^{2}$College of Intelligence and Computing, Tianjin University, Tianjin, 300072, China \\ 
 $^{3}$Department of Physics, University of Hong Kong, Hong Kong, China \\ 
 $^{\dag}$These authors contributed equally to this work
}

\begin{document}
\maketitle

\begin{abstract}
Recently, artificial intelligence has been extensively deployed across various scientific disciplines, optimizing and guiding the progression of experiments through the integration of abundant datasets, whilst continuously probing the vast theoretical space encapsulated within the data. Particularly, deep learning models, due to their end-to-end adaptive learning capabilities, are capable of autonomously learning intrinsic data features, thereby transcending the limitations of traditional experience to a certain extent. Here, we unveil previously unreported information characteristics pertaining to different frequencies emerged during our work on predicting the terahertz spectral modulation effects of metasurfaces based on AI-prediction. Moreover, we have substantiated that our proposed methodology of simply adding supplementary multi-frequency inputs to the existing dataset during the target spectral prediction process can significantly enhance the predictive accuracy of the network. This approach effectively optimizes the utilization of existing datasets and paves the way for interdisciplinary research and applications in artificial intelligence, chemistry, composite material design, biomedicine, and other fields.
\end{abstract}

After discovering this phenomenon, we have started conducting specific experiments. We also welcome researchers from other research groups to explore this issue together. We are actively seeking experimental collaborations on this phenomenon. If you are interested, please contact us at: \href{mailto: xing_xiaohua@tju.edu.cn}{xing\_xiaohua@tju.edu.cn},\href{mailto: wuliang@tju.edu.cn}{wuliang@tju.edu.cn}.

\section{Introduction}
The collection and analysis of data serve as the foundation for scientific understanding and discovery, constituting two central aims in science. However, the knowledge and laws encapsulated within are often too complex, surpassing the realm of human solvability. Even with the utilization of traditional numerical methods, it remains challenging to easily explore vast hypothesis spaces\cite{lecun2015deep,davies2021advancing}. The proliferation of artificial intelligence (AI) technology has transcended the conventional boundaries of isolated disciplines, enhancing the design and efficacy of data analysis and scientific research, thereby potentially reshaping scientific discoveries at various stages\cite{wang2022efficient}. Deep learning, as a significant approach in AI, employs an end-to-end learning methodology that not only circumvents the laborious process of manual feature extraction but also disregards the intricate underlying logic of model construction and learning. Consequently, since its inception, it has found extensive applications across diverse fields, attracting numerous researchers seeking solutions to distinct problems within their respective disciplines\cite{lynch2023multi,mankowitz2023faster,weiss2023deep,chen2023single,shaltout2019spatiotemporal,an2019deep,ma2018deep,zhou2020diagnostic,ma2019probabilistic,thieme2023deep,chen2023transformer}. In traditional photonics, similar challenges are encountered, as exemplified by the design of modulating devices for terahertz (THz) band. Due to the limited electromagnetic responses range of natural materials, metasurfaces have become the preferred choice in recent years\cite{han2020spectral,chen2022demand,zhao2018electromechanically,kim2022tunable,aieta2012aberration,liu2018generative,zhang2021genetic,li2017electromagnetic,ko2018wideband,pitchappa2015microelectromechanically}. However, the design process of metasurfaces faces numerous challenges. It requires specialized knowledge of Maxwell’s equations, determination of optimization starting points during experimental and simulation processes and incurs significant time costs due to continuous experiments or computations. Additionally, traditional simulation techniques such as finite-difference time-domain (FDTD) still heavily rely on researchers’ personal physical insights and intuitive reasoning, making the design of structures far from effortless. Furthermore, the relevant data involved in different studies are complex and difficult to further integrate and analyze, hindering the formation of generalized laws that are more amenable to widespread applications. 
Here, we adeptly address the intricate and non-intuitive relationship between metasurface structures and their optical responses using deep learning techniques. We have successfully achieved the prediction of THz spectra and reverse engineering of structure design by using multiple deep learning models\cite{Yu2023sleep,chu2022transformer,chen2023transformer1} (CNN, LSTM, GRU and Transformer, see details in the supplementary materials). In terms of data utilization, we propose a simple method of simply adding supplementary multi-frequency inputs to the existing dataset during the target spectrum prediction process. Furthermore, it is inspiring that previously unreported patterns in different frequency information have emerged in this process. Specifically, we found that the low-frequency portion of the metallic structure contains more informative content than the high-frequency portion. This work provides new insights for the integration of deep learning in disciplines such as optical neural chips, optical metrology, image classification, composite material design, and biomedical research \cite{zhou2023transformer,wang2022single,lee2022learning,zhang2021optical,zuo2022deep,ashtiani2022chip,zhu2022space,mourgias2022noise}. Moreover, the emerged laws in different frequency information in this work contribute to a deeper understanding of the essence of various materials, which is of significant importance for advancing fundamental physics research \cite{shastri2021photonics,chen2023dispersion,grudinina2023collective,hashimoto2017all}.

\section{Results}
\label{Results}
\subsection{Metamaterial Structure}
\begin{figure}[htbp]
\centering
\includegraphics[scale=0.15]{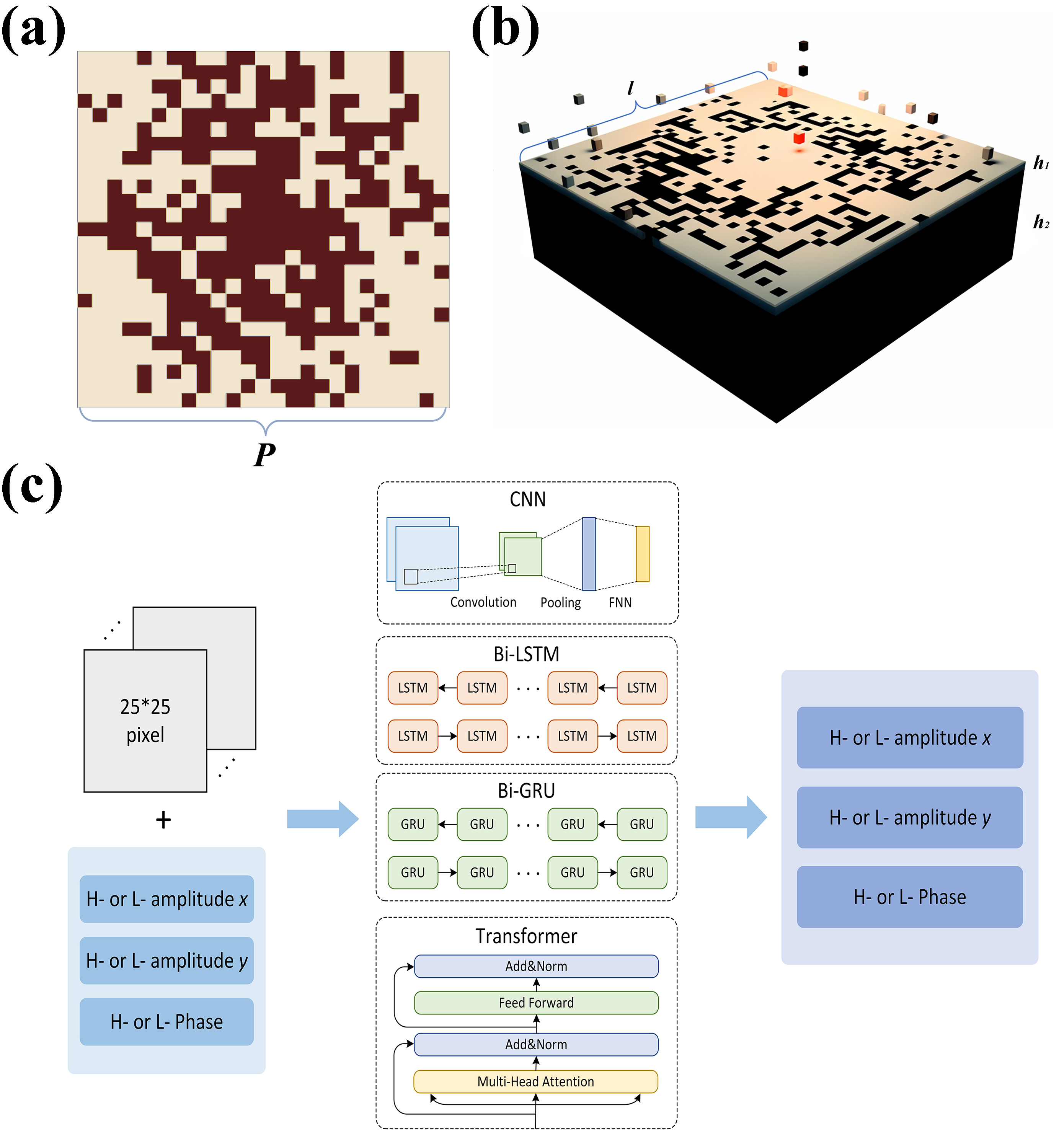}
\caption{\textbf{Schematic diagrams of metasurface structures and deep learning networks. a-b} Schematic diagrams of metasurface structures, where the brown sections represent ‘1’ pixels and the yellow sections represent ‘0’ pixels. \textbf{c} Schematic diagrams of CNN, LSTM, GRU, and Transformer networks. In forward design, the input of model is the 25*25 pixels matrix and the output is the optical responses, while the inverse design is reverse. ‘H’ and ‘L’ marked in the figure represent the relevant amplitude and phase data of high frequency and low frequency respectively.}
\label{fig1}
\end{figure}

In this work, we aim to construct a new ‘screen’ with sufficient structural diversity to achieve this goal (see details in the supplementary materials). We have chosen a 25*25 pixels screen as a single unit of the metasurface structure, which was subsequently arranged in a periodic array, as shown in Fig. \ref{fig1}. For each such unit, its substrate is composed of high- resistance silicon material with a side length of $\SI{200}{\upmu m}$ and a thickness of $\SI{500}{\upmu m}$. The ‘1’ unit on the screen is an aluminum flake with a side length of $\SI{8}{\upmu m}$ and a thickness of 200 nm, while the ‘0’ unit has no metal patch. Different arrangements of ‘0’ and ‘1’ units form different structural patterns. Although more complex screens could increase the freedom of pattern design, the current screen can already generate $2^{625}$ different pattern types. Further increases would only increase the computational difficulty of prediction, with little practical benefit.

\subsection{Forward and inverse design}
\begin{figure}[htbp]
\centering
\includegraphics[scale=0.15]{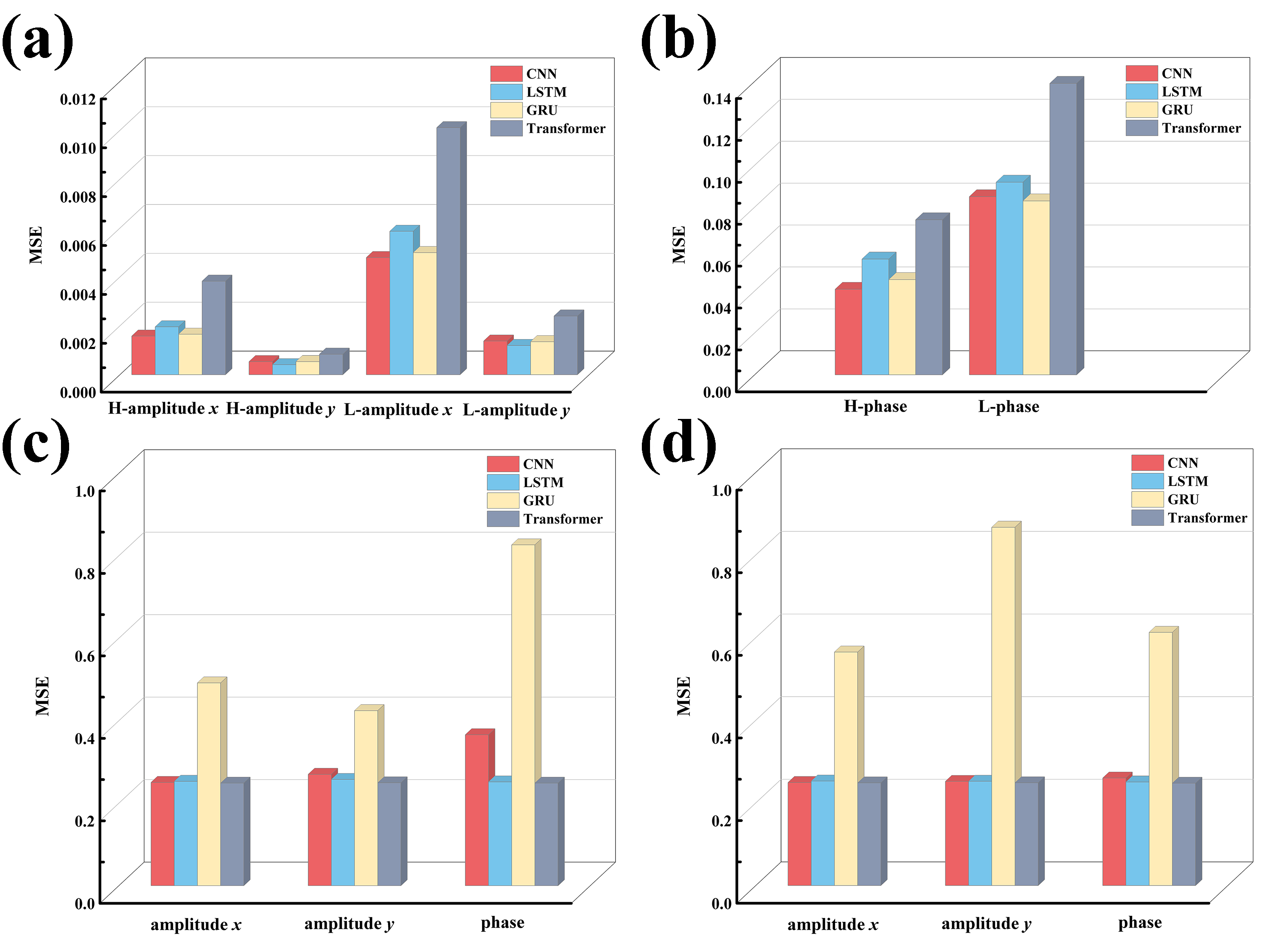}
\caption{\textbf{The prediction performance of different models on high-frequency, low-frequency data and inverse network. a} Prediction performance of different models on high-frequency and low-frequency amplitudes. \textbf{b} Model performance comparison on the prediction of high-frequency and low-frequency phases. \textbf{c} Effectiveness of predicting structural parameters using high-frequency data. \textbf{d} Effectiveness of predicting structural parameters using low-frequency data.}
\label{fig2}
\end{figure}

No matter it is material property prediction, genetic sequence analysis, or structural design, addressing such issues often requires extensive data analysis. However, the complex underlying logic and profound interdisciplinary knowledge required make the acquisition of such data less straightforward. Thus, deep learning, an important branch of machine learning, has become a new approach to solving such problems. The adaptive learning process from the input to the output and the powerful computing capacity have alleviated many engineering hitches that were previously difficult to analyze. In this work, the complex design of a metasurface screen also adopts this solution, effectively reducing various limitations brought by traditional simulation techniques and greatly saving design time and costs. For the metasurface screen, the metal patches on top mainly play a role in modulating terahertz waves. Such a structure has a thin and uniform thickness in the propagation direction, therefore, this type of screen can be approximately regarded as a two-dimensional pattern. The two-dimensional pattern formed by the distribution of different pixels corresponds one-to-one with its target spectral response. Therefore, for the analysis of such case, the first model that comes to mind is the classic model in the CV field, CNN.
We have tried two different approaches for data construction. The first approach treats the entire metasurface screen as a complete image, while the second approach represents it as a sentence with 25 tokens (the detailed information can be found in the first part of the ‘Method’ section.). In this model’s operating mode, it is essential to learn when to pass information, and conventional RNN networks are prone to problems such as gradient disappearance and state missing bound. Therefore, in this work, we used LSTM and GRU models to solve the above problems and obtained the forward network result shown in Fig. \ref{fig2}. Because different scenarios require different target response working frequency bands, here we further refine the entire frequency range analysis into low-frequency (0-1 THz) and high-frequency (1-2 THz) parts. Under the premise of the incident polarization state being x-polarization, we analyze the transmitted x-polarization and orthogonal y-polarization amplitude and phase of the metamaterial screen. Specific information on forward design can be found in the supplementary materials.
Transformer model has been extensively utilized as a backbone network in CV and NLP due to its remarkable performance. Here we also present the performance results of the Transformer model in forward prediction (Fig. \ref{fig2}). The performance of Transformer in the forward network is not as good as the other three models. This is because the unique attention mechanism of Transformer is mainly used to solve sequence data tasks with long-range dependencies, while the forward network is more like a structural analysis task, which can typically be replaced by simulation software. However, in the design of the inverse network with long sequences as input, the role of Transformer is very prominent.
Because we do not need to consider specific optical response design for the time being, we randomly input high-frequency data (Fig. \ref{fig2}c) and low-frequency data (Fig. \ref{fig2}d) to inversely predict the pixel distribution of the metamaterial screen. Regardless of whether it is high-frequency or low-frequency data, Transformer performs the best, and the MSE is generally around 0.249. This excellent analysis result is mainly due to its attention mechanism. Unlike CNN, this attention mechanism is not local, and can capture long-range dependencies between different positions, providing more precise overall control of the ‘semantic’ of the entire device and closer to the expected functional design.

\subsection{The optimization effect of multi-frequency supplementary input}
After successfully establishing the forward and inverse networks, we further attempted to improve the model’s analytical capability. In order to avoid incurring additional costs associated with training new datasets, we plan to design a scheme aimed at improving the utilization of existing datasets. Inspired by our dual processing approach to the pattern and text of the metasurface screen, we consider that the analysis of photonic target responses typically involves specifying the target working frequency such as the specific amplitude and phase at around 0.8 THz, which is less concerned about the performance at other frequencies such as 1.8 THz. We propose to boost the non-working frequency band data as input on the basis of the original metasurface structure parameters to improve the model’s prediction accuracy. 

\begin{figure}[htbp]
\centering
\includegraphics[scale=0.15]{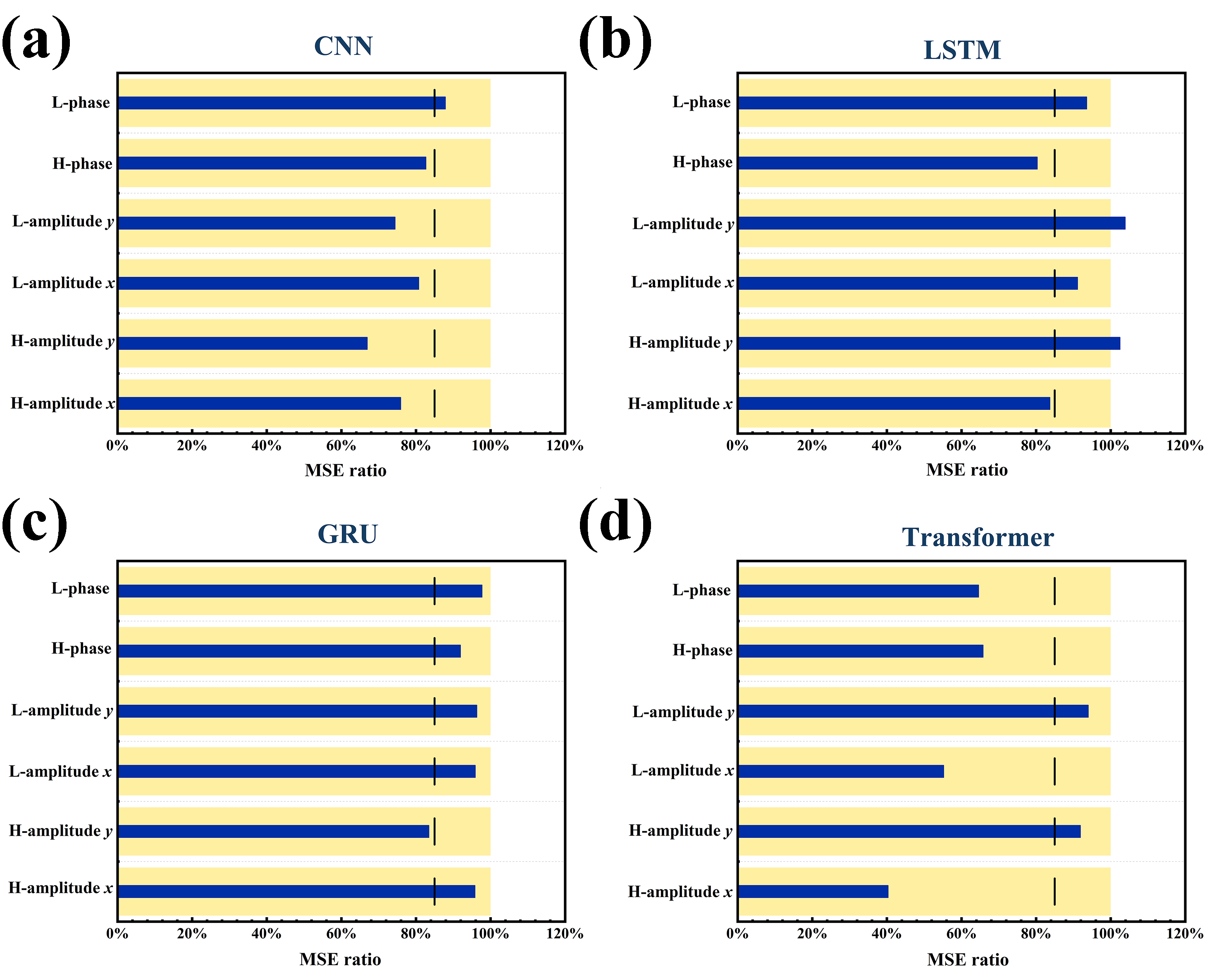}
\caption{\textbf{Comparison of prediction accuracy after applying optimization methods. a-d} The accuracy of CNN, LSTM, GRU, and Transformer models in predicting the amplitude and phase data of high-frequency and low-frequency transmission states with different polarizations is influenced by the introduction of additional multi-frequency data supplementation. We used 100$\%$ as the baseline criterion for evaluating optimization effectiveness (yellow region) and set 85$\%$ as the threshold for distinguishing significantly improved optimization performance (black vertical line) when comparing the reduction in MSE after optimization (blue bars). It is evident that after undergoing the optimization process, there is a noticeable decrease in MSE across the majority of the dataset.}
\label{fig3}
\end{figure}

To demonstrate the wide applicability of this approach, we assume that the target working frequency is either high or low and compare the effects of direct prediction results with low-frequency or high-frequency supplementary inputs, as shown in Fig. \ref{fig3}. After introducing the other half of the frequencies as supplementary input, the MSEs of different models are effectively reduced. Among them, the CNN model has the best overall prediction effect, and can still reduce the MSE by about 20$\%$-30$\%$ on the basis of high-precision prediction. The Transformer model has the best improvement effect on its own, specifically in the prediction of the amplitude of high-frequency transmission of x-polarization state, with an MSE of only about 40$\%$ of the previous value after adding low-frequency data. To intuitively demonstrate the model’s prediction effect after adding multi-frequency supplementary inputs, we selected some schematic diagrams of the forward networks.

\begin{figure}[htbp]
\centering
\includegraphics[scale=0.15]{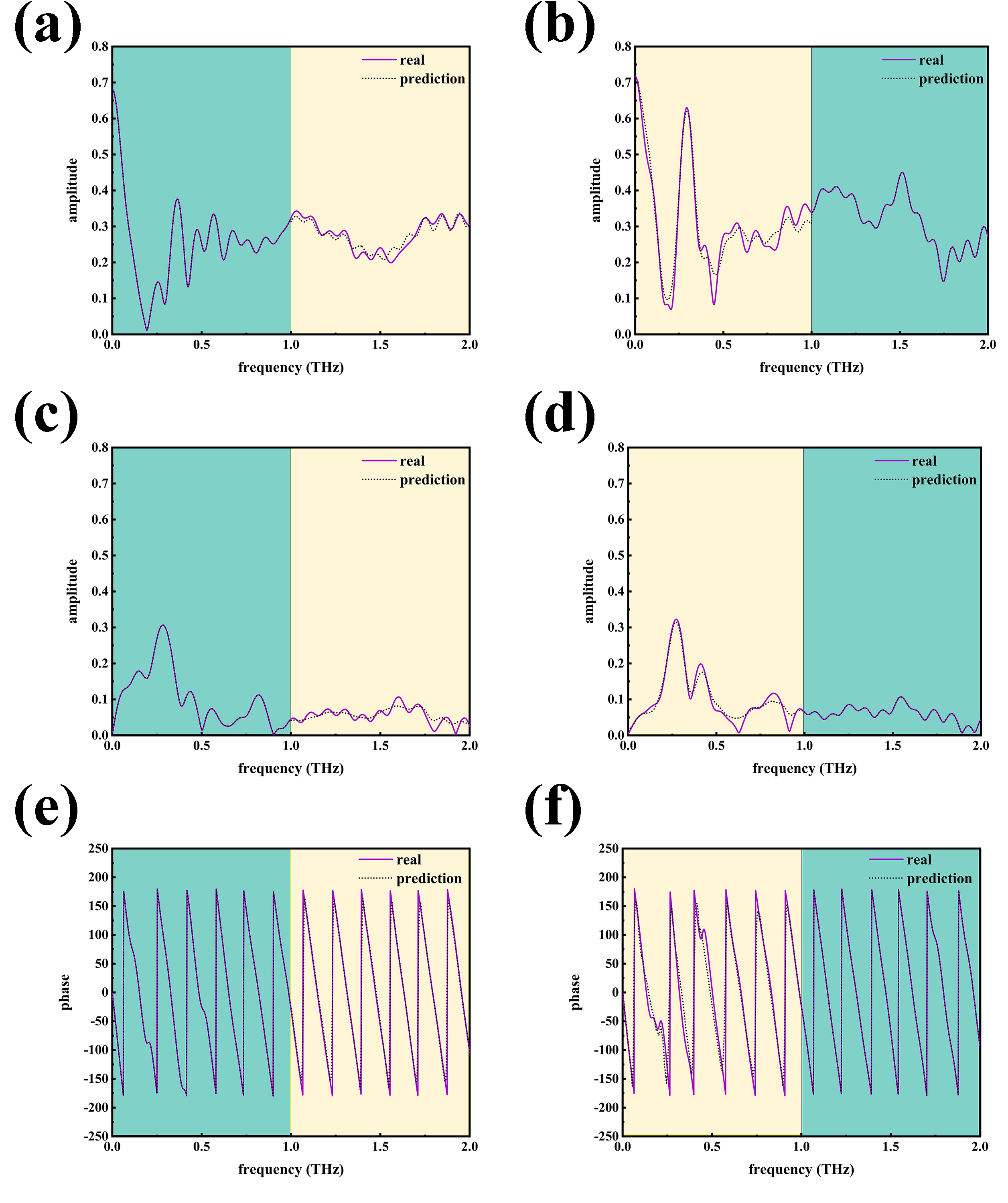}
\caption{\textbf{Schematic diagrams of the prediction performance for high-frequency and low-frequency data after optimization. a-b} We compare the real data (purple solid line) with the predicted data (black dashed line) to demonstrate the varying prediction performance of the optimized network model across different frequency ranges. The green region represents the frequency information data used as supplementary input, while the yellow region represents the area for validating the optimized prediction performance. While a and b represent the prediction results for the high-frequency and low-frequency amplitudes of the x-polarized state. \textbf{c-d} Prediction results for the high-frequency and low-frequency amplitudes of the y-polarized state. \textbf{e-f} Prediction results for the high-frequency and low-frequency phases.}
\label{fig4}
\end{figure}

As shown in Fig. \ref{fig4}, the yellow area is the frequency range to be analyzed, the green area is the supplemented known input, the purple solid line represents the true data distribution, and the black dashed line represents the predicted data distribution. Figs. \ref{fig4}a and \ref{fig4}b show the effect of predicting the other half of the x-polarization state amplitude after adding low-frequency and high-frequency data inputs, respectively, while Figs. \ref{fig4}c and \ref{fig4}d show the effect of predicting the other half of the orthogonal y-polarization state amplitude after adding low-frequency and high-frequency data inputs. Figs. \ref{fig4}e and \ref{fig4}f show the effect of predicting the other half of the phase after adding low-frequency and high-frequency data inputs. It shows that regardless of the parameters, good prediction effects can be attained. Furthermore, as mentioned earlier, we did not choose labeling or other forms of human intervention in this work and tried to let the intelligent agent model maintain an adaptive learning approach as much as possible. This design concept did bring some encouraging results beyond our expectations, which had not been mentioned in previous physical experiences. 

\subsection{Emergence of frequency information non-uniform distribution}
Particularly, we found that the low-frequency portion of the metallic structure contains more informative content than the high-frequency portion. Traditional dispersion analysis is often highly constrained, necessitating the fulfillment of numerous limiting conditions(see details in the supplementary materials). Based on our experimental results (Fig. \ref{fig2} and Fig. \ref{fig3}) under the same input condition, the low-frequency data predicting high-frequency data process has an obvious higher accuracy than the high-frequency data predicting low-frequency data process, with a difference of almost four times in accuracy. Because the number of dispersion relations contained in our screen is as many as $2^{625}$, which is approximately infinite in statistics, we may speculate that the effective information contained in the low-frequency portion should be greater than that in the high-frequency portion. 
The supplementation of additional frequencies to the original dataset has shown an improvement in the network’s predictive accuracy\cite{bell1995program,modinos2001applications,chin1994green,li1998transmission,pendry1996extremely,zhou2005saturation}. We attribute this effect to a similar principle observed in previous complex frequency excitation designs\cite{guan2023overcoming}, which enhance the overall information representation and reduce information errors by transitioning from single-frequency excitation to complex frequency excitation. This transition enables practical applications such as highly sensitive sensing. Moreover, regarding the emergence that the low-frequency portion contains more information compared to the high-frequency portion, we attribute this pattern to the minimal overall absorption of the constructed metasurface screen substrate. Consequently, the metallic portion on top plays a predominant role in this process. In the traditional understanding of optics, high-frequency and low-frequency information data are often believed to differ only in their frequency bands, with their information content assumed to be uniformly distributed and equivalent. However, through comparative analysis in this study, we propose the hypothesis that high and low-frequency information may not actually exhibit uniform distributions. The reasons for their uneven information distribution, we suggest, can be analyzed using the Drude model. According to Drude’s model, the dielectric properties of metals can be expressed through complex permittivity, which typically comprises a real part and an imaginary part (dielectric loss):
\begin{equation}
\varepsilon(\omega)=1-\frac{\omega_{p}^{2}}{\omega^{2}+\gamma^{2}}+\mathrm{i} \frac{\omega_{p}^{2} \gamma}{\omega\left(\omega^{2}+\gamma^{2}\right)}
\label{eq1}
\end{equation}

Where $\omega_{p}=\sqrt{n e^{2} / \varepsilon_{0} m^{*}}$ is the plasmon frequency, with n the electron density and m* the effective electron mass, $\gamma$ is a measure of the inelastic scattering rate, which is strongly dependent on the film thickness. At THz frequencies which are far below the plasma frequency of metals (usually lies in ultraviolet frequency range), from equation (\ref{eq1}), with the increase of frequency, the imaginary part (dielectric loss) of permittivity also increases, indicating a higher loss of the metasurface. Therefore, the high-frequency portion contains less information than the low-frequency portion. This has important guiding significance for the speculation and derivation of the dispersion functional relationship of generalized causality systems. Moreover, the proposed method can also be applied to the latest work on complex frequency information compensation\cite{guan2023overcoming,archambault2012superlens}. By selecting the target frequency range more accurately and supplementing information from other frequency ranges, it is possible to effectively enhance efficiency.

\section{Conclusion}
In summary, we have uncovered hitherto unreported non-uniform information distribution related to different frequencies in our work on predicting the THz spectral responses of metasurfaces based on deep learning, specifically that the low-frequency section of the THz spectrum in metal structures contains more information content than the high-frequency section. With respect to the prediction work itself, we have, for the first time, proposed a scheme to augment the accuracy of network predictions and enhance the utility of existing datasets by adding supplementary multi-frequency inputs into the current dataset during the spectral prediction process. In terms of dataset construction, we have diverged from traditional three-dimensional geometric modeling approaches and referred to the concept of masking language modeling to the field of optics. With respect to model selection, we have explored the effects of various types of deep learning models on spectral prediction, successfully constructing forward and inverse networks for amplitude and phase parameters, covering a wide frequency range with over a thousand sampling points, whilst maintaining high prediction accuracy. This design approach offers fresh perspectives for the intersection of deep learning with other domains and underscores the potential of AI to unlock scientific discoveries that were previously unattainable, suggesting that even mature research fields can still give rise to new physical conclusions, which holds positive implications for further in-depth analysis of fundamental physics.

\section*{Method}
\subsection*{1. Neural Networks}
Treating the entire screen as an image, the CNN model is trained to learn the correspondence between all screen images in the dataset and their target spectra as much as possible. While this approach has achieved good predictive results, it has a drawback. That is, although these screen images contain almost all the information of the metasurface structure, the data acquired is intertwined. All the underlying physical mechanisms regarding metasurface structures are encapsulated and stored within a comprehensive image matrix that encompasses a vast array of 2625 possibilities. On the basis of a limited dataset, this presentation of information is not conducive to deep learning model. Of course, we can consider labeling these screen images, but such labels will actually bring the analysis and learning back to previous traditional experience, which cannot meet the target of constructing an adaptive learning process automatically. Accordingly, we began to explore the segmentation of the overall information of such screens by treating each column of 25 pixels blocks as a token. This approach establishes a potential relationship between different tokens similar to the contextual semantics of a sentence. As a result, the input is not a complete image but rather a sequence of 25 tokens comprising ‘01’ that conveys the semantic information of target spectral response. For such text processing, the RNN models in the NLP field are better suited which process the next token based on the summarized information, which is repeated throughout the process.
We utilized four mainstream neural networks for forward and inverse prediction: CNN, GRU, LSTM, and Transformer. CNN consists of convolutional layers, pooling layers, and fully connected layers, performing convolutional calculations by sliding filters to extract local features. LSTM and GRU are commonly used for processing sequential data to address the issue of long-term dependencies in vanilla RNN model. LSTM introduces three gate mechanisms (forget gate, input gate, and output gate) to control the flow of information in memory cells, preserving earlier information. GRU is a simplified version of LSTM, combining the forget and input gates into an update gate to achieve similar effects with fewer parameters. Transformer's key component is the attention mechanism, which extracts global dependencies in a sequence and assigns higher attention weights to important information. In this paper, we used the self-attention mechanism, where queries, keys, and values are the hidden states from the previous layer. The training objective for all models is to minimize the mean square error (MSE) loss function, which measures the matching degree between predicted values and actual values.

\subsection*{2. Experimental details}
To demonstrate the generalization performance of the models, our experimental results were obtained through 5-fold cross-validation. In all experiments, the training set was trained for 20 epochs with a batch size of 8. The optimizer used was Adam, with $\beta_{1}$ set to 0.9 and $\beta_{2}$ set to 0.999. The learning rate was set to $1e^{-3}$. The neural networks were constructed using the open-source machine learning framework PyTorch. All experiments were conducted on a single Nvidia a6000 GPU.

\section*{Disclosures}
The authors declare no conflicts of interest.

\section*{Acknowledgments}
Xiaohua Xing and Yuqi Ren contributed equally to this work. The authors acknowledge the financial support from the National Key Research and Development Program of China (Grant No. 2022YFA1203502), and the National Natural Science Foundation of China (NSAF, No. U2230114).

\section*{Authors contributions}
L.Wu initiated the project. X.H.Xing and Y. Q. Ren constructured the neural network model and performed the experiments. D.Zou, Q.K. Zhang and B. X. Mao prepared the figures. X.H.Xing and L.Wu developed the theoretical model and prepared the manuscript. S. Zhang review and edit the manuscript. D. Y. Xiong and J.Q.Yao supervised the project.

\section*{Data and code availability}
The data and code that support the plots within this paper and other findings of this study are available from the corresponding authors upon reasonable request.

\bibliographystyle{unsrt}  
\bibliography{references}  

\end{document}


\maketitle

\tableofcontents

\newpage
\section*{Design details for different models}
\phantomsection\addcontentsline{toc}{section}{Design details for different models}\tolerance=500
In selecting models, we utilize different deep learning models, such as Convolutional Neural Networks (CNN), Long Short-Term Memory (LSTM), and Gated Recurrent Unit (GRU), which have demonstrated outstanding performance in computer vision (CV) and natural language processing (NLP) domains. Furthermore, we introduce the Transformer model, which has exhibited exceptional performance in large language models. Its attention mechanism enables capturing long-range dependencies and focusing on key information relevant to the prediction target, thus enhancing the accuracy of learning and analyzing optical target parameters. 
We have validated the significant impact of this approach on improving the accuracy of network predictions. Unlike traditional three-dimensional geometric parameter modeling approaches, our data set construction approach innovatively draws inspiration from the masking language modeling concept used in the analysis of molecular and biological functionalities. We treat different units as analogous to letters forming words and sentences, thereby defining the meaning of the document. By exploring the learning effectiveness of the image matrix information or sequences composed of 25 tokens (where different token sequences represent different spectral responses) obtained through different modeling approaches, we successfully construct forward and inverse networks for amplitude and phase parameters, covering a wide frequency range with over a thousand sampling points while maintaining high prediction accuracy. Furthermore, we have also attempted to process the information of the metasurface structure as images. All of this can be achieved using a simple low-resolution (25*25) metasurface ‘screen’.

\section*{Details of metamaterial structural design}
\phantomsection\addcontentsline{toc}{section}{Details of metamaterial structural design}\tolerance=500
If a metasurface structure takes the form of a certain fixed geometric shape (as shown in the following figure), modifying its structural parameters typically only alters its performance under that specific modulation type but scarcely affects its overall modulation effect on amplitude. For instance, it is challenging to transform its narrowband modulation effect into Fano resonance or broadband bandpass solely by altering the length and width of a particular structure. To demonstrate the universality of the proposed scheme in optics and other fields, the selected metamaterial structure needs to be capable of generating as much optical response types as possible. Traditional structures such as split-ring resonators cannot meet such requirements. Therefore, in this work, we aim to construct a new ‘screen’ with sufficient structural diversity to achieve this goal.

\begin{figure}[htbp]
\centering
\includegraphics[scale=0.1]{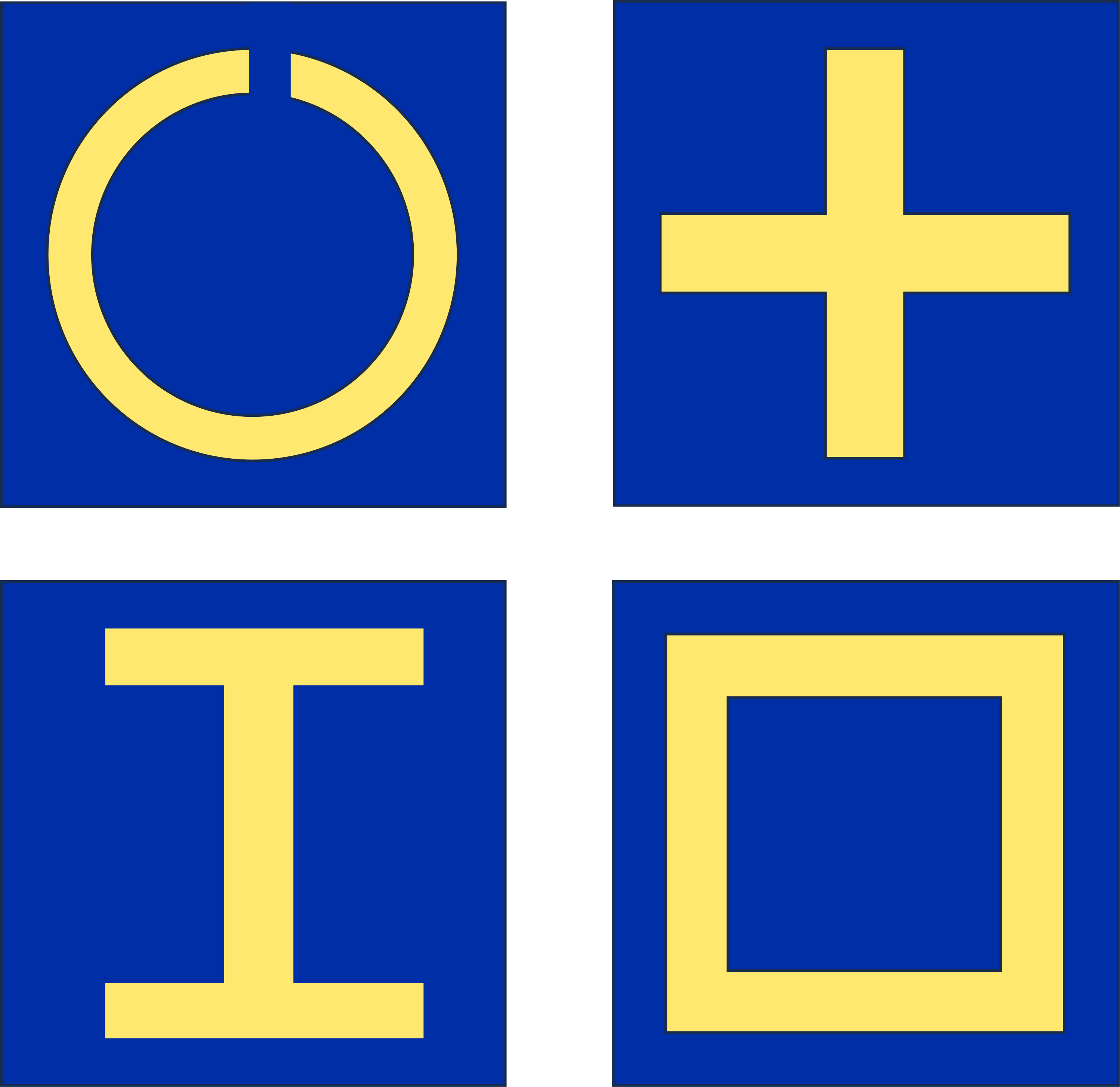}
\caption{Schematic diagram of metasurface structures with classic deterministic shapes.}
\label{sup}
\end{figure}

\newpage
\section*{Forward design effect details}
\phantomsection\addcontentsline{toc}{section}{Forward design effect details}\tolerance=500
In forward design, Fig. 2a is the low-frequency and high-frequency analysis of the amplitude modulation effect generated by the structure, and Fig. 2b is that of the phase modulation effect. The prediction results of all the networks (CNN, LSTM, or GRU) are quite accurate. The mean square error (MSE) of the predicted amplitudes for both high and low frequency parts are lower than that of the phase prediction. This is because the phase analyzed here is a variable of the cycle oscillation, and the sharp mutation will cause the increase of predicted MSE to a certain extent, but the overall MSE is still relatively low, effectively meeting the expectations of the forward network design. Among them, for both high and low-frequency amplitude predictions, the CNN model predicts best, with MSEs of 0.00158 and 0.00479, respectively. For the prediction of the y-polarization amplitude, the LSTM model predicts both high and low-frequency data best, with MSEs of 0.00041 and 0.0012, respectively. For the prediction of the phase, the most suitable model is different. The CNN model performs best for high-frequency data, with an MSE of 0.04083, while the GRU model performs best for low-frequency data, with an MSE of 0.08299. From the above results, it can be seen that treating the constructed metamaterial screen as both a complete image and a statement with specific semantics has different advantages in different parameter analyses.

\section*{Some examples of traditional dispersion analysis methods}
\phantomsection\addcontentsline{toc}{section}{Some examples of traditional dispersion analysis methods}\tolerance=500
The dispersion relation of different frequencies analyzed in this work can be regarded as a causal reasoning in the physical system, where the external input of a signal or some kind of action leads to a modulated output effect after the action rule is applied through some functional relationship. This connection is referred to as the dispersion relation. The functional relationship is usually very difficult to analyze. For conventional photonic crystals, we can start from Maxwell’s equations and obtain their dispersion relations with respect to the characteristic frequency given a wave vector k. For specific shapes, such as cylindrical or spherical scattering system units, we can also use multiple scattering methods or the Layer-KKR method for analysis. However, traditional analysis methods are not effective for the strong frequency dispersion of metasurfaces. For specific structures, such as the double-notch resonant annular metamaterial structure, people have attempted to use LC equivalent circuit models for analysis, but these methods are clearly not applicable to the more complex optical responses of the metasurface screen designed in this work or even more extensive dispersion relations.